\pdfoutput=1

\documentclass[11pt]{article}

\usepackage[]{emnlp2021}

\usepackage{times}
\usepackage{latexsym}

\usepackage[T1]{fontenc}

\usepackage[utf8]{inputenc}

\usepackage{microtype}
\usepackage{adjustbox}
\usepackage{booktabs}
\usepackage{multirow}
\usepackage{amsmath}
\usepackage{amssymb}
\usepackage{tabularx}

\def\figref#1{Fig.~\ref{#1}}
\def\secref#1{Sec.~\ref{#1}}
\def\tabref#1{Table~\ref{#1}}
\def\eqnref#1{Eqn.~\ref{#1}}

\title{Learning and Analyzing Generation Order for \\Undirected Sequence Models}

\author{Yichen Jiang \and Mohit Bansal \\
UNC Chapel Hill \\
  \texttt{\{yichenj, mbansal\}@cs.unc.edu} \\
}
\begin{document}
\maketitle

\begin{abstract}
Undirected neural sequence models have achieved performance competitive with the state-of-the-art directed sequence models that generate monotonically from left to right in machine translation tasks.
In this work, we train a policy that learns the generation order for a pre-trained, undirected translation model via reinforcement learning.
We show that the translations decoded by our learned orders achieve higher BLEU scores than the outputs decoded from left to right or decoded by the learned order from~\citet{mansimov2019generalized} on the WMT'14 German-English translation task.
On examples with a maximum source and target length of 30 from De-En, WMT'16 English-Romanian, and WMT'21 English-Chinese translation tasks, 
our learned order outperforms all heuristic generation orders on four out of six tasks.
We next carefully analyze the learned order patterns via qualitative and quantitative analysis. 
We show that our policy generally follows an outer-to-inner order, predicting the left-most and right-most positions first, and then moving toward the middle while skipping less important words at the beginning.
Furthermore, the policy usually predicts positions for a single syntactic constituent structure in consecutive steps.
We believe our findings could provide more insights on the mechanism of undirected generation models and encourage further research in this direction.\footnote{Our code is publicly available at \\ \url{https://github.com/jiangycTarheel/undirected-generation}}
\end{abstract}

\section{Introduction}
Directed neural sequence models such as the Transformer~\cite{vaswani2017attention} with causal self-attention masks have been widely used for language generation tasks like machine translation and summarization.
In such models, each token only depends on the left context and thus they can be naturally applied in the monotonic left-to-right generation scheme.
On the other hand, in the undirected sequence models such as the Transformer without the causal mask, each token depends on the full left and right context.
The application of an undirected sequence model in language generation is not straightforward, as it can simply peek into the future for the target token if trained in the same way as a directed model.
To circumvent this problem, previous works~\cite{ghazvininejad-etal-2019-mask,mansimov2019generalized} trained undirected Transformer models using the masked language modeling (MLM)~\cite{devlin-etal-2019-bert} objective with the masking ratio randomly varying in the range of (0\%, 100\%].
At test time, the model first predicts the length $T$ of the sequence to be generated and begins with a sequence of $T$ mask tokens.
Then, at each step, the model selects one or more positions to update, masks the original tokens at those positions, and predicts the target tokens to replace the original ones there.
However, how to select this order of tokens to update throughout the process still remains an open and interesting question for the research community.

In this work, we study this question in a focused setting where the model can only replace one mask token per step for $T$ steps.\footnote{In this setting, the model cannot re-mask and re-generate an already-predicted token.}
Under this linear-time decoding setup, we can easily compare the generation model's behaviors when decoding in different orders.
Previously, \citet{mansimov2019generalized} designed and bench-marked various heuristic orders under this setting.
They further built and trained a policy network with reinforcement learning (RL) to predict the target generation order for the pre-trained masked language model dynamically for each example.
We aim to learn better generation orders by improving upon the previous RL objective. 
First, instead of using the change in the edit distance between the partially generated sequence and the ground-truth sequence as the reward function for the order policy, 
we directly use the BLEU scores of the ultimate sequence generated in the order sampled from the policy network as the reward because edit distance is too rigid and evaluates a sequence as individual tokens instead of n-grams, and thus cannot reflect the actual quality of a generated sequence order. 
Furthermore, on the De$\xrightarrow{}$En task, \textit{the learned policy by~\citet{mansimov2019generalized} either follows the left-to-right order or generates the final punctuation first and then proceeds from left to right in around 75\% of the generations.}
Therefore, this might indicate that the policy converges prematurely to the left-to-right order which is a strong local optimum.
And hence, to encourage the policy to better explore the entire action space instead of putting all weights on one position at every step too early, we add a negative entropy penalty to the policy's output distribution.
When sampling an order during the training, a larger entropy means that the policy can explore and evaluate more possible orders and hence lead to a better policy ultimately.

On the WMT14' English$\leftrightarrow$German translation tasks, using the same pre-trained undirected Transformer fixed during the learning of the policy, the sequences decoded in the orders predicted by our policy receive higher BLEU scores 
than the outputs decoded in the order learned in the previous work~\cite{mansimov2019generalized}.
On the De$\xrightarrow{}$En task, our learned order also beats the Left2Right baseline on the test set.
On the WMT16' Romanian$\xrightarrow{}$English task, our learned order again outperforms the Left2Right order and is competitive with the best heuristic Easy-First order.
Moreover, our learned policy performs strongly when trained and evaluated on examples with the source and target shorter than 31 
tokens, outperforming all heuristic generation orders (Uniform, Left2Right, Easy-First, Least2Most) on the dev sets of WMT'14 En$\leftrightarrow$De, WMT'16 Ro$\xrightarrow{}$En, and WMT'21 Zh$\xrightarrow{}$En tasks.
Therefore, while our learned policy achieves the best overall scores on more than 2/3 of all En$\leftrightarrow$De examples and 1/2 of all Ro$\xrightarrow{}$En examples, it is expectedly harder for the policy to learn an optimal order on the remaining very long sequences, which is an interesting direction for future works. 

Next, we focus on a detailed and important qualitative analysis (and human study) to reveal that, interestingly, our policy in general follows an outer-to-inner order, predicting the left-most and right-most positions first and move toward the middle.
In \tabref{table:learned_order_eg_teargas}, we show that when decoding from left to right, the Transformer model generates two ``gas" at the end that makes the sentence ungrammatical.
This error is due to the poor space planning from the model: when the decoding proceeds until ``using" is sampled, the model finds that there are three spaces (masks) left but only two tokens (``gas .") are actually needed.
Instead, when decoding by the policy's order, the last three tokens (``tear gas .") are generated within the first eight steps, therefore avoiding the space issue early in the generation.
We also show that our learned outer-to-inner order outperforms several heuristic ones with fixed strides.
Additionally, we perform a loss function ablation study to understand that both the BLEU reward and the entropy penalty contribute to our learned order's convergence to an outer-to-inner pattern.
Furthermore, as shown in~\tabref{table:learned_order_eg_affleck}, the policy tends to skip some less important tokens (e.g., determiner, suffix) at first and come back to them at the last few steps.
This helps the planning of the entire sequence as more important words are generated earlier and can provide more useful context information in the following de-masking steps.
Finally, we observe that the learned policy usually predicts positions for a single syntactic constituent structure in consecutive steps.
We believe our quantitative results and qualitative findings could provide more insights on the mechanism of undirected generation models.
\section{Background}
\paragraph{Directed Sequence Generation Models.}
Most generation models nowadays are directed as they are trained and decode exclusively from left to right.
The widely-used Transformer~\cite{vaswani2017attention} decoder employs a causal self-attention layer to mask out the attention to the future tokens during the training. 
At the test time, the decoder takes the previously generated tokens as the input and predicts the next token from left to right.

\paragraph{Undirected Sequence Generation Models.}
Other than training the auto-regressive Transformer model with causal self-attention in a monotonic direction, previous works~\cite{ghazvininejad-etal-2019-mask,mansimov2019generalized} have tried to enable a decoder to generate tokens in non-monotonic orders. 
We follow the training procedure on the En$\leftrightarrow$De task from~\citet{mansimov2019generalized}, where they first initialize from a Transformer-based cross-lingual language model~\cite{lample2019XLM} pre-trained on the large monolingual corpus in both English and German.
They then train the Transformer as a conditional masked language model (MLM)~\cite{devlin-etal-2019-bert}, where it is trained to predict some randomly masked tokens from the target sequence, given the complete source sequence and other non-masked tokens of the target sequence. The input is constructed by concatenating the German and English sequences separated by a special token. At each iteration, the sentence in one language is randomly selected as the source and the corresponding sentence in the other language is the target and is masked in random positions.
Instead of masking the tokens with a fixed probability (e.g., 0.15 in BERT~\cite{devlin-etal-2019-bert}), they randomly vary the masking masking probability from (0\%, 100\%] in order to mimic all possible partially generated sequences the model might face during the test time:
at the beginning, the model sees a fully masked sequence (100\% masks) and predicts the token at a selected position; at the final step, there is only one remaining mask (close to 0\%) and the model replace it with a predicted token.

\paragraph{Decoding MLM with Pre-defined Order.}
Because the target sequence tokens are randomly masked, the model is trained to generate in all possible orders and therefore is undirected.
Hence, a generation order is needed at test time in order to decode this conditional masked language model auto-regressively (token by token).
Previous works~\cite{ghazvininejad-etal-2019-mask,mansimov2019generalized} have explored a range of heuristic generation orders, including random order (Uniform), left-to-right, first replacing the least-likely masked token (Least-to-Most), and first replacing the masked token where the model is the most certain (judging by the entropy of the model's output distribution) about the actual token (Easy-First), etc.
It was shown that the Easy-First performs the best in greedy search and the Left-to-Right order outperforms the other in beam search. 
\textit{All three deterministic heuristic orders achieved significantly higher BLEU scores than the random order.}
The results suggest that, even though the Transformer model is trained to follow all possible generation orders, \textit{it still prefer some orders over the others to achieve higher BLEU scores at the test time.}
Therefore, there remains a question: is there an optimal order for the MLM Transformer in achieving the highest BLEU scores, possibly beating all these pre-defined, heuristic orders?

\begin{figure*}[t!]
\vskip -0.1in
\begin{center} 
\includegraphics[width=0.9\textwidth]{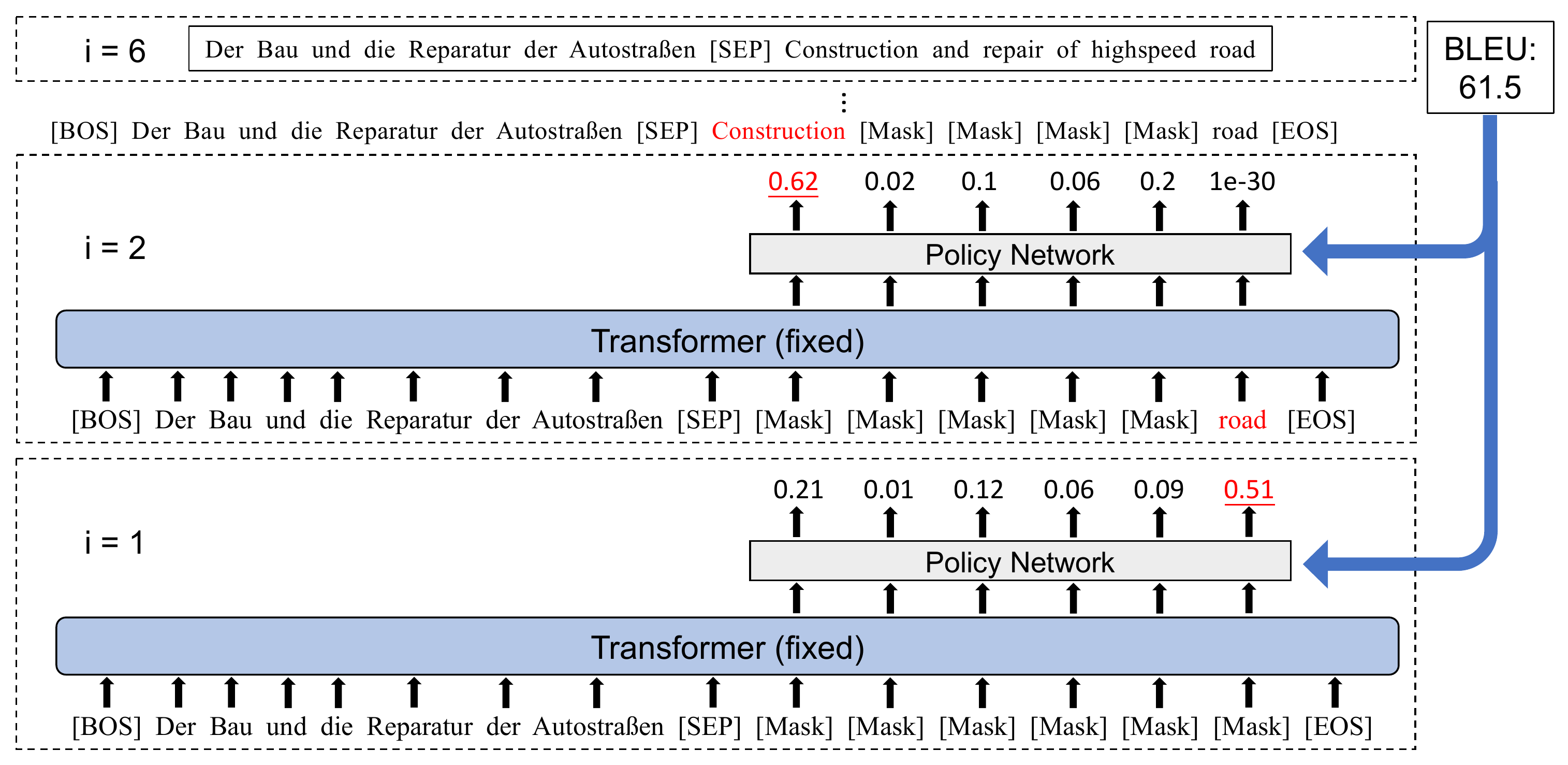}\end{center} 
\vskip -0.15in
\caption{A visualization of our model and the RL procedure adapted from~\citet{mansimov2019generalized}.
At every step, the policy network samples a masked position and the Transformer model generates a token for that position. 
At the end, the BLEU score for the fully-generated translation is fed back to the policy network as the reward.
\vspace{-8pt}}
\label{fig:model}
\end{figure*}

\paragraph{Decoding MLM with Predicted Order.}
In the quest for generation orders that adapt to the trained masked language model optimally, \citet{mansimov2019generalized} trained a policy that predicts the position of the next masked token to be replaced with reinforcement learning (RL).
Specifically, they use the change in edit distance between the partial sequence and the ground-truth sequence after substituting a masked token as the reward function, and optimize the policy with the proximal policy optimization (PPO)~\cite{schulman2017proximal}.
The order learned by their RL policy underperforms the Left2Right order on the En$\xrightarrow{}$De translation task in both greedy and beam search.
On the De$\xrightarrow{}$En translation task, the learned order marginally outperforms the Left2Right order but falls short of the Easy-First order in the greedy search scenario.
Most importantly, on the De$\xrightarrow{}$En task, \textit{in around 75\% of the generations, the learned policy either follows the left-to-right order or generates the final punctuation first and then proceeds from left to right.}
Therefore, we think this policy converges to a local optimum because Left2Right is already a strong order that beats the starting point (a random order).
For this reason, in our work, we aim to train the policy to better explore the action space by improving the training objective.
We show that by following the generation order predicted by our policy, the translations generated by the masked-language-model Transformer receive higher BLEU scores than following the previous learned or heuristic~\cite{mansimov2019generalized} orders;
and our learned policy predicts a wide variety of orders that deviate from the left-to-right baseline.

\section{Methods}
In this section, we introduce our policy network (\secref{ssec:policy_net}) that predicts the generation order for a pre-trained conditional MLM and our reinforcement learning (RL) objective (\secref{ssec:RL_objective}).
We visualize our models and the RL procedure in~\figref{fig:model}.

\subsection{Predicting Generation Order}
\label{ssec:policy_net}
Same as the setup in~\citet{mansimov2019generalized}, we use a pre-trained, dual-lingual, conditional masked language model Transformer as the generation model and freeze its parameters during the following training.
To learn the generation order for this model, we follow~\citet{mansimov2019generalized} to construct two separate networks, a policy network and a value network, which are two separate 2-layer MLPs added on top of the Transformer's output of the last layer.
The policy network projects the output vectors at all positions to scalar logits, 
which are fed to the softmax to produce the distribution $\mathbf{P}_{pol}$ over all masked positions.

\subsection{Reinforcement Learning Objectives}
\label{ssec:RL_objective}
We use the Advantage Actor-Critic Algorithm (A2C) to optimize the order policy. 
The model starts with a target sequence of all masks as the input.
At every step $i$, we sample one position $p_i$ with a mask from $\mathbf{P}_{pol}$ and replace it with the greedily generated token from the Transformer model.
The updated target sequence is then fed to the model at the next step.
For the example in~\figref{fig:model}, at step 1 we first sample the right-most position from the policy, where the generation model predicts the word ``road" to replace the mask.
At step 2, the left-most position is sampled from the policy and the model predicts ``Construction" at this position.
When the entire target translation is generated, we calculate its BLEU score as the overall reward $R$ for the sampled order $[p_1, p_2, ..., p_T]$.
At every step $i$, the value network also outputs the critic $v_i$. 
The advantage for the step $i$ is then calculated as: $A_i = \gamma^{T-i}R - v_i$, where $\gamma$ is the discount factor.
The RL loss function is: $\mathcal{L}_{RL} = -\frac{1}{T}\sum_{i=1}^{T} \mathrm{log}(\mathbf{P}_{pol}(p_i)) A_i $.

\paragraph{Entropy Penalty.}
In order to encourage our policy to better explore the action space, we additionally minimize the negative entropy of the policy's output probability $\mathbf{P}_{pol}$. The final loss function is the weighted sum of the two losses:
\begin{equation} \label{eq:entropy_loss}
\begin{split}
\mathcal{L}_{ent} &= -\frac{1}{T^2}\sum_{i=1}^{T}\sum_{j=1}^{T}\mathbf{P}_{pol}(p_i^j)\mathrm{log}(\mathbf{P}_{pol}(p_i^j)) \\
\mathcal{L}_{total} &= \mathcal{L}_{RL} + \lambda \mathcal{L}_{ent}
\end{split}
\end{equation}
where $\lambda$ is the entropy loss coefficient.
When sampling an order during the RL training, a larger entropy means that the policy can better explore and evaluate more possible orders and hence lead to a better policy ultimately.
Empirically, we found that the model converges quickly to a left-to-right order without this negative entropy penalty.
\section{Experiments}

\begin{table*}[t]
\centering
\begin{small}
\begin{tabular}[t]{cc||ccccc|c}
\toprule
 Task &  & Uniform & Left2Right & Easy-First & Least2Most & Learned & Learned (Ours) \\
\midrule
\multirow{2}{*}{De$\xrightarrow{}$En} & 
test  & 26.01 & 28.34 & \textbf{29.00} & 28.85 & 28.47 & 28.64 \\
& dev & 25.70 & \textbf{28.21} & \textbf{28.19} & 27.30 & - & 27.92\\
\midrule
\multirow{2}{*}{En$\xrightarrow{}$De} & 
test  & 21.01 & \textbf{24.27} & 23.73 & 23.08 & 24.10 & \textbf{24.19} \\
& dev & 20.75 & \textbf{23.60} & 23.27 & 22.61 & - & \textbf{23.66} \\
\midrule
\multirow{2}{*}{Ro$\xrightarrow{}$En} & 
test  & 30.62 & 31.82 & \textbf{32.47} & 31.60 & - & 32.18 \\
& dev & 30.95 & 32.79 & \textbf{33.22} & 32.69 & - & \textbf{33.23} \\
\midrule
\multirow{2}{*}{En$\xrightarrow{}$Ro} & 
test  & 32.75 & \textbf{35.55} & 35.18 & 34.59 & - & 35.38\\
& dev & 33.43 & \textbf{36.38} & \textbf{36.34} & 35.14 & - & 36.13 \\
\bottomrule
\end{tabular}
\vspace{-5pt}
\caption{Results (BLEU) on the test sets of WMT'14 En$\leftrightarrow$De and WMT'16 En$\leftrightarrow$Ro translation tasks.
The model is decoded with greedy search and different decoding orders. The heuristic and learned orders are from~\citet{mansimov2019generalized} while the last column is the results of our learned orders. 
}
\label{table:full_test}

\end{small}
\end{table*}

\subsection{Experimental Setup}

\paragraph{Datasets.} We train and evaluate our order policy on WMT'14 English-German and WMT'16 English-Romanian translation datasets.
The former contains 4.5M sentence pairs and the latter contains 2.8M sentence pairs.
More details of our experimental setup and hyperparaters are discussed in supplementary.

\subsection{Heuristic Order Baselines}
\label{ssec:heuristic_order}
To demonstrate the strength of our learned order policy, we compare it against four heuristic orders introduced in previous works~\cite{ghazvininejad-etal-2019-mask,mansimov2019generalized}.
\paragraph{Uniform.} We randomly samples a masked position at every step.
\paragraph{Left2Right.} We select the left-most masked position at every step.
\paragraph{Least2Most.} At the step $i$, we pass the current input sequence to the generation model and, for each position $j$, evaluate how unlikely the current token $y_j^i$ is under the source $X$ and current target: $\phi_{\text{logp}} = - logp(y_j = y_j^i | y^i_{<j}, \langle mask\rangle,y^i_{>j}, X)$. We then select and replace the mask at the position with the largest $\phi_{\text{logp}}$.
\paragraph{Easy-First.} Other than $\phi_{\text{logp}}$, we further consider the negative entropy of the generation model's output distribution at every position: $\phi_{\text{negent}}=-\mathcal{H}(y_j^{i+1} | y^i_{<j}, \langle mask\rangle,y^i_{>j}, X)$
Intuitively, we want to replace the mask with a new word that the model is highly certain of (low entropy).
We then select and replace the mask at the position with the largest $\alpha_{\text{logp}}\cdot\phi_{\text{logp}}+\alpha_{\text{negent}}\cdot\phi_{\text{negent}}$.\footnote{We set $\alpha_{\text{logp}}=1$, $\alpha_{\text{negent}}=0.9$ for En$\xrightarrow{}$De and $\alpha_{\text{logp}}=1$, $\alpha_{\text{negent}}=1$ for the other three tasks.}

\begin{table*}[t]
\centering
\begin{small}
\begin{tabular}[t]{cc||cccc|c}
\toprule
 & max length & Uniform & Left2Right & Easy-First & Least2Most & Learned (Ours) \\
\midrule
\multirow{2}{*}{De$\xrightarrow{}$En} & 
20   & 27.22 & 28.48 & 28.96 & 28.14 & \textbf{29.57} \\
& 30 & 26.90 & 28.89 & 29.21 & 28.12 & \textbf{29.40} \\
\midrule
\multirow{2}{*}{En$\xrightarrow{}$De} & 
  20 & 22.50 & 23.83 & 24.06 & 23.08 & \textbf{24.32} \\
& 30 & 21.67 & 23.86 & 23.80 & 22.88 & \textbf{24.02} \\
\midrule
\multirow{2}{*}{Ro$\xrightarrow{}$En} & 
  20 & 30.53 & 30.63 & 30.92 & 30.89 & \textbf{31.09}\\
& 30 & 30.14 & 31.26 & \textbf{31.49} & 31.27 & \textbf{31.46}\\
\midrule
\multirow{2}{*}{En$\xrightarrow{}$Ro} & 
  20 & 33.68 & 34.74 & \textbf{35.67} & 34.37 & 34.65 \\
& 30 & 33.09 & 34.77 & \textbf{35.22} & 33.96 & 34.88\\
\midrule
\midrule
\multirow{2}{*}{Zh$\xrightarrow{}$En} & 
 20 & 19.94 & 19.54 & 20.10 & 18.37 & \textbf{22.18} \\
& 30 & 17.68 & 19.44 & 18.81 & 18.03 & \textbf{19.54} \\
\midrule
\multirow{2}{*}{En$\xrightarrow{}$Zh} & 
 20   & 27.25 & 28.31 & 26.69 & \textbf{28.72} & 28.61 \\
& 30 & 28.54 & 29.69 & 28.83 & \textbf{30.37} & 29.83 \\
\bottomrule
\end{tabular}
\caption{Results (BLEU) on the partial dev sets with both source and target with a maximum length of 20/30, from WMT'14 En$\leftrightarrow$De, WMT'16 En$\leftrightarrow$Ro, and WMT'21 En$\leftrightarrow$Zh translation tasks. 
}
\label{table:dev_less30}
\end{small}
\end{table*}

\begin{table}[t]
\centering
\begin{small}
\begin{tabular}[t]{c|ccccc}
\toprule
 Task / $\lambda$ & $0.01$ & $0.001$ & $0.0005$ & $0.0001$ & 0\\
\midrule
De$\xrightarrow{}$En & 
 28.56 & \textbf{29.40} & 28.98 & 28.74 & 28.46\\
Ro$\xrightarrow{}$En & 
 31.41 & \textbf{31.46} & \textbf{31.46} & 31.20 & 30.92\\
\bottomrule
\end{tabular}
\caption{Ablation BLEU with different $\lambda$, evaluated on the dev sets with source\&target shorter than 31 tokens. 
\label{table:lambda_ablation}}
\vspace{-5pt}
\end{small}
\end{table}

\subsection{Main Results}
We present our evaluation results by BLEU~\cite{papineni-etal-2002-bleu} in~\tabref{table:full_test}.~\footnote{We can only report the test-set accuracy of the learned order in~\citet{mansimov2019generalized} for the En$\leftrightarrow$De as they did not release the code for learning the generation order.}
On the De$\xrightarrow{}$En test set, the Easy-First generation order achieves the highest BLEU scores among all orders, while the translations decoded by our learned orders receives higher BLEU scores compared to the ones decoded from left to right and the ones decoded by previous learned orders~\cite{mansimov2019generalized}.
On the En$\xrightarrow{}$De test set, the translations generated from left to right outperforms the translation decoded by any other heuristic or learned orders.
Here, our learned order again beats the previous learned order and is on par with the Left2Right baseline on the dev set.
For the Ro$\xrightarrow{}$En task, our learned order achieves the highest BLEU scores among all generation orders on the dev set and only falls behind the Easy-First on the test set.
On the En$\xrightarrow{}$Ro task, the Left2Right and Easy-First order marginally outperforms our learned order.
We calculate the statistical significance between our learned order and the Left2Right order on the full dev sets. For De$\xrightarrow{}$En, En$\xrightarrow{}$De, and En$\xrightarrow{}$Ro, the difference between the Left2Right heuristic and our learned order is insignificant. On the Ro$\xrightarrow{}$En task, our learned order achieves significant (p-value $<$ 0.05) improvement compared to the Left2Right heuristic. 
Therefore, our learned order achieves better or equal performances to the Left2Right on full dev sets.

We further compare our learned orders, which are trained on the training examples with a maximum source and target length of 30, against all heuristic orders on the dev-set examples where the source and target sequences have a maximum length of 20/30.
As shown in~\tabref{table:dev_less30}, in three (De$\leftrightarrow{}$En, Ro$\xrightarrow{}$En) out of four tasks, translations decoded by our learned orders achieve the highest BLEU scores compared to all heuristic orders.
This suggests that our order policy can learn effective generation orders better on upto medium and above-average length sequences (24.21 on avg. for De-En and 27.62 on avg. for Ro-En).
For the examples with very long source and target sentences, the task of learning an optimal generation order expectedly becomes more challenging.
To understand the lower performance there, we examine 100 examples with length 30 or more where our learned policy underperforms a left-to-right heuristic. 
One observation is that the model is still prone to making mistakes at the final steps when the remaining space is tight, similar to how the model makes errors when generating from left to right. 
This indicates that this fixed-length setting might be a bottleneck and we should look for a more flexible approach that allows the model to dynamically adjust the length when necessary.

\paragraph{Zh-En Translation Results.}
To complement the previous results, we further learn and analyze the generation order for translations of English$\leftrightarrow$Chinese, which are considered to be relatively more different in terms of grammar.
We present the automatic metric scores in the lower section of~\tabref{table:dev_less30}.
In Zh$\xrightarrow{}$En task, our learned generation order again beats all four heuristic orders on dev-set examples with source and target sequences shorter than 20/30 tokens.
In En$\xrightarrow{}$Zh task, our learned order outperforms the Left2Right and Easy-First heuristics, but outperformed by the Least2Most order.
We hope our work encourages future research into the generation order of more types of language pairs.

\subsection{The Effects of the Entropy Penalty}
We present an ablation study on different $\lambda$ values we used in \eqnref{eq:entropy_loss}. 
A larger coefficient $\lambda$ means a stronger penalty that encourages the policy's output distribution to have a larger entropy.
As shown in \tabref{table:lambda_ablation}, the best $\lambda$ value is 0.001 for the De$\xrightarrow{}$En task and 0.001/0.0005 for the Ro$\xrightarrow{}$En task.
We observe that with this negative entropy penalty added to the RL loss, the learned policy achieves significantly better BLEU scores than without the penalty ($\lambda$=0), suggesting that the penalty can indeed make the policy to better explore its action space instead of converging to a local minimum prematurely.

\subsection{Reward Function}
The experiments we present above are conducted by using the BLEU scores of the sequence greedily decoded by the sampled order.
We also explore using the overall likelihood of the generated sequence measured by the pre-trained undirected model, as the reward.
This approach avoids the hand-crafted BLEU metric and directly uses the undirected model as the source of critic.
As shown in \tabref{table:reward_ablation}, using BLEU as the reward function is slightly better than the likelihood reward.
Thus, we experiment with BLEU as the reward function consistently in this work.

\begin{table}[t]
\centering
\begin{small}
\begin{tabular}[t]{c|ccccc}
\toprule
 Reward / $\lambda$ & $0.01$ & $0.001$ & $0.0005$ & $0.0001$ \\
\midrule
BLEU & 
 28.56 & \textbf{29.40} & 28.98 & 28.74\\
Likelihood & 
 29.02 & 29.17 & 29.16 & 28.85 \\
\bottomrule
\end{tabular}
\caption{BLEU with different reward function and $\lambda$, evaluated on the De$\xrightarrow{}$En dev set with source\&target shorter than 31 tokens. 
\label{table:reward_ablation}}
\vspace{-5pt}
\end{small}
\end{table}

\begin{table*}[t]
\centering
\begin{small}
\begin{tabular}[t]{cc}
\toprule
\textbf{Step} & \textbf{Translation decoded by our learned order}\\
\midrule
(source) & Nu exista rapoarte privind eventuale raniri , insa unele persoane lacrimau, dupa utilizarea gazelor lacrimogene . \\
(target) & There are no reports of any injuries , but some people were crying after the use of teargas .\\
(L2R) & There are no reports of any injuries , but some people are crying foul after using tear gas gas . \\
\midrule
4 & [There are no reports] \underline{\hspace{2mm}} \underline{\hspace{2mm}} \underline{\hspace{2mm}} \underline{\hspace{2mm}} \underline{\hspace{2mm}} \underline{\hspace{2mm}} \underline{\hspace{2mm}} \underline{\hspace{2mm}} \underline{\hspace{2mm}} \underline{\hspace{2mm}} \underline{\hspace{2mm}} \underline{\hspace{2mm}} \underline{\hspace{2mm}} \underline{\hspace{2mm}} \underline{\hspace{2mm}} \underline{\hspace{2mm}} \\
6 & [There are no reports] \underline{\hspace{2mm}} \underline{\hspace{2mm}} \underline{\hspace{2mm}} \underline{\hspace{2mm}} \underline{\hspace{2mm}} \underline{\hspace{2mm}} \underline{\hspace{2mm}} \underline{\hspace{2mm}} \underline{\hspace{2mm}} \underline{\hspace{2mm}} \underline{\hspace{2mm}} \underline{\hspace{2mm}} \underline{\hspace{2mm}} \underline{\hspace{2mm}} \{gas .\} \\
8 & [There are no reports] of \underline{\hspace{2mm}} \underline{\hspace{2mm}} \underline{\hspace{2mm}} \underline{\hspace{2mm}} \underline{\hspace{2mm}} \underline{\hspace{2mm}} \underline{\hspace{2mm}} \underline{\hspace{2mm}} \underline{\hspace{2mm}} \underline{\hspace{2mm}} \underline{\hspace{2mm}} \underline{\hspace{2mm}} tear \{gas .\}\\
13 & [There are no reports] of [any injuries] \{, but\} some \underline{\hspace{2mm}} \underline{\hspace{2mm}} \underline{\hspace{2mm}} \underline{\hspace{2mm}} \underline{\hspace{2mm}} \underline{\hspace{2mm}} \underline{\hspace{2mm}} \{tear gas .\}\\
17 & [There are no reports] of [any injuries] \{, but\} some \underline{\hspace{2mm}} \underline{\hspace{2mm}} \underline{\hspace{2mm}} \{after the use of\} tear \{gas .\}\\
20 & [There are no reports] of [any injuries] \{, but\} some [people are crying] \{after the use of\} tear \{gas .\}\\
\bottomrule
\end{tabular}
\vspace{-5pt}
\caption{An Example of Ro$\xrightarrow{}$En translation generated by our learned order. The words in [] are generated in consecutive steps from left to right; the words in \{\} are generated in consecutive steps from right to left.
}
\label{table:learned_order_eg_teargas}
\end{small}
\end{table*}

\begin{table}[t]
\centering
\begin{small}
\begin{tabular}[t]{cc|ccc|c}
\toprule
\multirow{2}{*}{Task} & \multirow{2}{*}{Length} & \multicolumn{3}{c}{Outer-to-Inner} & \multirow{2}{*}{Learned} \\
 & & s=1 & s=2 & s=3 &   \\
\midrule
\multirow{2}{*}{De$\xrightarrow{}$En} & $<30$ & 28.73 & 28.86 & 28.55 & \textbf{29.40}\\
 & full & 27.86 & 27.81 & 27.78 & \textbf{27.92}\\
 \midrule
\multirow{2}{*}{En$\xrightarrow{}$De} & $<30$ & 23.36 & 23.26 & 23.19 & \textbf{24.02} \\
 & full & 22.97 & 22.91 & 23.01 & \textbf{23.66}\\
\midrule
\multirow{2}{*}{Ro$\xrightarrow{}$En} & $<30$ & 31.05 & 31.04 & 31.04 & \textbf{31.46} \\
 & full & 32.91 & 32.91 & 32.85 & \textbf{33.23}\\
 \midrule
\multirow{2}{*}{En$\xrightarrow{}$Ro} & $<30$ & 34.23 & 34.15 & 33.95 & \textbf{34.88}\\
 & full & 35.71 & 35.64 & 35.60 & \textbf{36.13}\\
\bottomrule
\end{tabular}
\caption{
Ablation comparison between learned order and heuristic outer-to-inner orders with different strides (1, 2, 3). Evaluated on the full dev sets of WMT'14 En$\leftrightarrow$De and WMT'16 En$\leftrightarrow$Ro translation tasks and partial dev sets with a maximum length of 30.
All models are decoded with greedy search.
}
\label{table:heuristic_outer2inner}

\end{small}
\end{table}

\begin{table*}[t]
\centering
\begin{small}
\begin{tabular}[t]{cc}
\toprule
\textbf{Step} & \textbf{Translation decoded by our learned order}\\
\midrule
(source) & Gleichzeitig ist es Afflecks erster Film , der nicht im heimatlichen Boston des Regisseurs spielt . \\
(target) & It is also Aff@ leck 's first picture , which does not take place in the director 's hometown of Boston .\\
(L2R) & At the same time , Aff@ leck 's first film is not to be shown in Boston 's native director 's . \\
\midrule
8 & \underline{\hspace{2mm}} \underline{\hspace{2mm}} \underline{\hspace{2mm}} \underline{\hspace{2mm}} \underline{\hspace{2mm}} \underline{\hspace{2mm}} \underline{\hspace{2mm}} \underline{\hspace{2mm}} \underline{\hspace{2mm}} \underline{\hspace{2mm}} \underline{\hspace{2mm}} \underline{\hspace{2mm}} \underline{\hspace{2mm}} \underline{\hspace{2mm}}
\{place in the director 's native Boston .\} \\
13 & At \underline{\hspace{2mm}} [same time , Aff@] \underline{\hspace{2mm}} \underline{\hspace{2mm}} \underline{\hspace{2mm}} \underline{\hspace{2mm}} \underline{\hspace{2mm}} \underline{\hspace{2mm}} \underline{\hspace{2mm}} \underline{\hspace{2mm}}
\{place in the director 's native Boston .\}\\
15 & At \underline{\hspace{2mm}} [same time , Aff@] le@ \underline{\hspace{2mm}} 's \underline{\hspace{2mm}} \underline{\hspace{2mm}} \underline{\hspace{2mm}} \underline{\hspace{2mm}} \underline{\hspace{2mm}}
\{place in the director 's native Boston .\}\\
17 & At \underline{\hspace{2mm}} [same time , Aff@] le@ \underline{\hspace{2mm}} 's [first film] \underline{\hspace{2mm}} \underline{\hspace{2mm}} \underline{\hspace{2mm}}
\{place in the director 's native Boston .\}\\
19 & At the [same time , Aff@] le@ x 's [first film] \underline{\hspace{2mm}} \underline{\hspace{2mm}} \underline{\hspace{2mm}}
\{place in the director 's native Boston .\}\\
22 & At the [same time , Aff@] le@ x 's [first film] [does not take] \{place in the director 's native Boston .\}\\
\bottomrule
\end{tabular}
\caption{An Example of De$\xrightarrow{}$En translation generated by our learned order. The words in [] are generated in consecutive steps from left to right; the words in \{\} are generated in consecutive steps from right to left.
}
\label{table:learned_order_eg_affleck}
\end{small}
\end{table*}

\begin{table*}[t]
\centering
\begin{small}
\begin{tabular}[t]{cc}
\toprule
\textbf{Step} & \textbf{Translation decoded by our learned order}\\
\midrule
(source) & Doch alles wurde besser , als der " geliebte Vater " Heydar Aliyew das Ruder übernahm .
 \\
(target) & However , everything became better when the " beloved father " Hey@ dar Ali@ yev took the ru@ dder .
\\
(L2R) & But everything went well when Hey@ dar Ali@ ye@ v took over the " beloved father " of Ukraine . \\
\midrule
2 & \underline{\hspace{2mm}} \underline{\hspace{2mm}} \underline{\hspace{2mm}} \underline{\hspace{2mm}} \underline{\hspace{2mm}} \underline{\hspace{2mm}} \underline{\hspace{2mm}} \underline{\hspace{2mm}} \underline{\hspace{2mm}} \underline{\hspace{2mm}} \underline{\hspace{2mm}} \underline{\hspace{2mm}} \underline{\hspace{2mm}} \underline{\hspace{2mm}} \underline{\hspace{2mm}} \underline{\hspace{2mm}} \underline{\hspace{2mm}} \underline{\hspace{2mm}} \{over .\} \\
4 & \underline{\hspace{2mm}} [everything went] \underline{\hspace{2mm}} \underline{\hspace{2mm}} \underline{\hspace{2mm}} \underline{\hspace{2mm}} \underline{\hspace{2mm}} \underline{\hspace{2mm}} \underline{\hspace{2mm}} \underline{\hspace{2mm}} \underline{\hspace{2mm}} \underline{\hspace{2mm}} \underline{\hspace{2mm}} \underline{\hspace{2mm}} \underline{\hspace{2mm}} \underline{\hspace{2mm}} \underline{\hspace{2mm}} \{over .\} \\
7 & But [everything went] [well when] \underline{\hspace{2mm}} \underline{\hspace{2mm}} \underline{\hspace{2mm}} \underline{\hspace{2mm}} \underline{\hspace{2mm}} \underline{\hspace{2mm}} \underline{\hspace{2mm}} \underline{\hspace{2mm}} \underline{\hspace{2mm}} \underline{\hspace{2mm}} \underline{\hspace{2mm}} \underline{\hspace{2mm}} \underline{\hspace{2mm}} \{over .\} \\
11 & But [everything went] [well when] \underline{\hspace{2mm}} [" beloved father "] \underline{\hspace{2mm}} \underline{\hspace{2mm}} \underline{\hspace{2mm}} \underline{\hspace{2mm}} \underline{\hspace{2mm}} \underline{\hspace{2mm}} \underline{\hspace{2mm}} \underline{\hspace{2mm}} \{over .\} \\
12 & But [everything went] [well when] the [" beloved father "] \underline{\hspace{2mm}} \underline{\hspace{2mm}} \underline{\hspace{2mm}} \underline{\hspace{2mm}} \underline{\hspace{2mm}} \underline{\hspace{2mm}} \underline{\hspace{2mm}} \underline{\hspace{2mm}} \{over .\} \\
18 & But [everything went] [well when] the [" beloved father "] [Hey@ dar Ali@ ye@ v took] \underline{\hspace{2mm}} \underline{\hspace{2mm}} \{over .\} \\
20 & But [everything went] [well when] the [" beloved father "] [Hey@ dar Ali@ ye@ v took] \{the lead\} \{over .\} \\
\bottomrule
\end{tabular}
\caption{A De$\xrightarrow{}$En example where our learned order runs from left to right mostly. The words in [] are generated in consecutive steps from left to right; the words in \{\} are generated in consecutive steps from right to left.
 \vspace{-5pt}
}
\label{table:learned_order_eg_l2r}
\end{small}
\end{table*}

\section{Analysis Studies}
Generation using undirected sequence model has been an under-explored area compared to the conventional directed generation.
There are few studies that analyze the mechanism of this process and its difference with directed sequence generation.
Hence, to shed some light on the order preferences developed by the model when trained with conditional masked language modeling,
we present multiple qualitative and quantitative analyses on our learned generation orders and try to understand what causes the model to adapt such orders.

\subsection{Human Study on Learned Generation Order}
We randomly sample 100 examples from the dev sets of De$\xrightarrow{}$En and Ro$\xrightarrow{}$En respectively and summarize our order-pattern findings here.

\textbf{1. The learned policy starts from the left-most and right-most positions, and gradually moves toward the middle.}
As shown in \tabref{table:learned_order_eg_teargas}, the order policy first predicts the four head (left-most) positions at the first four steps and then predicts the three tail (right-most) positions in the reversed order in the next four steps.
After the head is extended till position 10 (``some") and the tokens ``after the use of" are prepended to the tail, the last three positions in the middle are filled and the entire sentence is generated.
On the other hand, when decoding from left to right, the Transformer model generates two ``gas" at the end that makes the sentence ungrammatical.
This error is due to the poor space planning from the model: when the decoding proceeds until ``using" is sampled, the model finds that there are three spaces (masks) remaining but only two tokens (``gas .") are needed.
However, when decoding by the policy's order, the last three tokens (``tear gas .") are generated within the first eight steps and thus prevents the incompatibility between the space and generation. 
Among the 200 examples analyzed, we identify 195 examples that follow this outer-to-inner order.
The remaining 5 examples follow either a mostly left-to-right (as shown in~\tabref{table:learned_order_eg_l2r}) or right-to-left generation order.
For this study, we additionally sample 100 generations from En$\xrightarrow{}$De and En$\xrightarrow{}$Ro each.
We observe the same outer-to-inner order in 86\% of the samples.
Therefore, we believe this learned \textbf{outer-to-inner} order is a preference first intrinsically developed by the Transformer during the masked language modeling training.
It is then effectively extracted by our order policy during reinforcement learning.

To demonstrate the advantage of the learned order that can flexibly control the generation process,
we implement several outer-to-middle baselines with fixed strides of 1, 2, 3. 
A stride of 2 means that the left-most 2 tokens will be generated first, then the right-most 2 tokens will be generated followed by the 3rd and 4th tokens from the left, and so on. 
As shown in~\tabref{table:heuristic_outer2inner}, among all four translation tasks, neither of the outer-to-middle heuristics can achieve better BLEU scores than our learned order. 
On De$\xrightarrow{}$En, the heuristic achieves a best of 28.86 BLEU (versus 29.4 from our learned policy) among three strides on the dev set of length < 30; 
and 27.81 (versus our 27.92) on the full dev set. 
On Ro$\xrightarrow{}$En, the heuristic achieves a best of 31.05 (versus our 31.46) on the dev set of length < 30 and 32.91 (versus our 33.23) on the full dev set. 
The same trend holds on En$\xrightarrow{}$De and En$\xrightarrow{}$Ro tasks. 
Therefore, we believe that it is crucial that our learned policy can execute the outer-to-middle order in a more flexible way. 
For example, our learned policy sometimes fills in a long sub-sequence consecutively before jumping to the other end.

\textbf{2. The learned policy sometimes skips the determiner and suffix tokens at first and predicts them at the end.}
For example in \tabref{table:learned_order_eg_affleck}, when generating the sentence, the 2nd position (``the") and the 8-th position (suffix ``x") are ignored by the policy when their surrounding contexts are generated, until the 5-th last and 4-th last steps.
We argue that by delaying the generation of such less important tokens to the end, the model can better plan the overall structure of the sentence by generating the key parts first. 
Instead, when generating from left to right, the model could face space issues at the end as shown in the example in~\tabref{table:learned_order_eg_teargas}.
To further solidify this argument, we automatically analyze 100 random samples each for De$\xrightarrow{}$En and Ro$\xrightarrow{}$En dev sets, where every sample has at least one suffix. We find that in 33\% of De$\xrightarrow{}$En examples and 56\% of Ro$\xrightarrow{}$En examples, a suffix is filled in later than both of its left and right neighbors, which means it is skipped at the first.
Therefore, we conclude that it is a frequent behavior of our model to skip suffix at first and predict it later.

\textbf{3. The learned policy predicts $n$ consecutive positions in $n$ consecutive steps to generate a constituent structure,} instead of frequently jumping back and forth between the head and tail.
Consider the example in \tabref{table:learned_order_eg_teargas}, where the first sub-sentence is decoded except for the preposition phrase (\texttt{PP}: ``of any injuries") after four iterations.
The positions and tokens predicted at the 7-th, 9-th and 10-th steps then complete this \texttt{PP}.
For the second sub-sentence, the noun phrase ``tear gas" is predicted first at iteration 6-8.
The entire preposition phrase (\texttt{PP}: ``after the use of tear gas") is then completed from the 14-th to 17-th steps consecutively.
Finally, the model fills in the subject noun phrase ``Some people" and the verb phrase ``are crying [\texttt{PP}]" to finish the generation process.
We quantitatively evaluate this claim by annotating the same 200 random samples. 
We manually inspect each example as we find that automatic matching can’t provide a straightforward estimate and sporadic parser errors further add on to the difficulty. 
As a result, we observe that among the 195 examples following an outer-to-inner order, 172 of them have at least one constituent structure finished in consecutive steps.

\subsection{Ablation of Cause of Outer-to-Inner Order}

As we observe that our learned order converges to an outer-to-middle trend while the learned order in~\citet{mansimov2019generalized} prefers a left-to-right order, we want to understand whether this is solely because of the entropy penalty we added, or the shift from edit-distance to BLEU reward also plays a part in this change of behavior.
We conduct the ablation study and find that when we use BLEU reward with no entropy penalty, the learned order quickly converges to an almost left-to-right policy. 
When we use entropy penalty but with edit-distance metric, the learned order converges to a combination of left-to-right and outer-to-inner policy, but with lower BLEU scores (27.68 for full De$\xrightarrow{}$En dev set vs 27.92 from our policy). 
Therefore, we believe it is the combination of BLEU and entropy penalty that makes the model fully explore the action space and finally converge to the outer-to-inner pattern.

\section{Related Works}

\paragraph{Undirected Generation with Iterative Refinement in Token Space.}
\citet{wang-cho-2019-bert} also explored approaches for generating text from a masked language model (MLM), such as BERT~\cite{devlin-etal-2019-bert}, by seeing it as a Markov random field language model and samples one token at a time.
Their setting differs from ours in that they used a pretrained MLM for unconditional language modeling.
Other than replacing one token at a time that result in a linear-time generation scheme, multiple previous works have tried to replace more than one token at time in a constant-time, non-autoregressive generation scheme.
\citet{lee-etal-2018-deterministic} proposed a model that replaces the tokens at all positions and keeps refining the previous outputs for multiple iteration.
\citet{ghazvininejad-etal-2019-mask} extended on this thread and introduced Mask-Predict that keeps replacing the tokens with low likelihood.
They showed that, after 10 iterations, the quality of the generated translations is competitive with the conventional autoregressive models on the WMT'14 En$\leftrightarrow$De and WMT'16 En$\leftrightarrow$Ro tasks.
\citet{liao-etal-2020-probabilistically} also investigated generation using masked language models and proposed a probabilistic masking scheme (PMLM).
\citet{wang-etal-2018-semi-autoregressive} proposed a semi-autoregressive generation scheme by predicting a consecutive chunk of tokens in parallel and repeats until the entire sequence is predicted.
\citet{kreutzer-etal-2020-inference} studied the Mask-Predict process in a similar semi-autoregressive setup and identified a thresholding strategy to improve upon the previous heuristics.
\citet{mansimov2019generalized} proposed a generalized framework for generating using masked language models by casting the generation as a Gibbs sampling process.
Under this framework, they proposed a log-linear model with different features ($\phi_{\text{logp}}$, $\phi_{\text{negent}}$ explained in \secref{ssec:heuristic_order}) for non-uniform position selection at every step.
They also trained a RL policy that selects one position to update at every step.
We build upon the same RL setup and further improve their objectives to achieve better BLEU scores as well as explore different and more effective generation orders.

\paragraph{Undirected Generation with Iterative Refinement in Continuous Vector Space.}
Other than iteratively refining the output tokens from the previous pass, another line of work used continuous latent variables and the distribution of the target sentence can be factorized over time given these variables~\cite{ma-etal-2019-flowseq,shu2020latent}.
\citet{lee-etal-2020-iterative} further improve the speed and performance of the EM-like inference procedure by training an inference network using the latent variable only.
Most recently, \citet{gu2020fully} improved the single-pass, fully non-autoregressive models by reducing the dependency in the output space.

\paragraph{Insertion-based Generation with Arbitrary Orders.}
Another generation scheme that is closely related to our work is the insertion-based generation~\cite{gu-etal-2019-insertion,pmlr-v97-stern19a}.
They also decode one token at a time within a linear-time generation scheme, and the insertion order is either some human-designed pre-defined order (e.g., left-to-right, balanced-tree, etc.) or a searched adaptive order found via beam search.
Recently, \citet{zhang-etal-2020-pointer} pre-trained an insertion-based model to generate text under specified lexical constraints.
Similar to the monotonic left-to-right generation, the insertion-based model also operates autoregressively and the length of the output is dynamically decided by predicting an end-of-sentence token.
We instead opt for using a masked language model and the output sequence length is fixed before the decoding starts. 
We further learn the generation order using Reinforcement Learning instead of relying on pre-defined heuristics.

\section{Conclusion}

In this work, we train a policy network with reinforcement learning to predict the generation order for an undirected sequence models.
It outperforms the Left2Right orders and previous learned orders on the full-length De$\leftrightarrow$En task.
When trained and evaluated on examples with source and target sequences of length 30 or less in De$\leftrightarrow$En, Ro$\leftrightarrow$En, and Zh$\leftrightarrow$En tasks,
our learned order outperforms all heuristic generation orders on four out of six tasks.
We show that our learned policy follows an outer-to-inner order and skips some less important words at first.
Moreover, it usually completes an entire constituent in consecutive steps. 
We hope our results and analyses could provide insights on undirected generation models and encourage future works on this topic.

\section*{Acknowledgements}
We thank the reviewers for their helpful comments. This work was supported by NSF-CAREER Award 1846185,  DARPA YFA17-D17AP00022, and ONR Grant N00014-18-1-2871. The views are those of the authors and not of the funding agency.
\bibliography{anthology,custom}

\begin{thebibliography}{22}
\expandafter\ifx\csname natexlab\endcsname\relax\def\natexlab#1{#1}\fi

\bibitem[{Devlin et~al.(2019)Devlin, Chang, Lee, and
  Toutanova}]{devlin-etal-2019-bert}
Jacob Devlin, Ming-Wei Chang, Kenton Lee, and Kristina Toutanova. 2019.
\newblock \href {https://doi.org/10.18653/v1/N19-1423} {{BERT}: Pre-training of
  deep bidirectional transformers for language understanding}.
\newblock In \emph{Proceedings of the 2019 Conference of the North {A}merican
  Chapter of the Association for Computational Linguistics: Human Language
  Technologies, Volume 1 (Long and Short Papers)}, pages 4171--4186,
  Minneapolis, Minnesota. Association for Computational Linguistics.

\bibitem[{Ghazvininejad et~al.(2019)Ghazvininejad, Levy, Liu, and
  Zettlemoyer}]{ghazvininejad-etal-2019-mask}
Marjan Ghazvininejad, Omer Levy, Yinhan Liu, and Luke Zettlemoyer. 2019.
\newblock \href {https://doi.org/10.18653/v1/D19-1633} {Mask-predict: Parallel
  decoding of conditional masked language models}.
\newblock In \emph{Proceedings of the 2019 Conference on Empirical Methods in
  Natural Language Processing and the 9th International Joint Conference on
  Natural Language Processing (EMNLP-IJCNLP)}, pages 6112--6121, Hong Kong,
  China. Association for Computational Linguistics.

\bibitem[{Gu and Kong(2020)}]{gu2020fully}
Jiatao Gu and Xiang Kong. 2020.
\newblock Fully non-autoregressive neural machine translation: Tricks of the
  trade.
\newblock \emph{arXiv preprint arXiv:2012.15833}.

\bibitem[{Gu et~al.(2019)Gu, Liu, and Cho}]{gu-etal-2019-insertion}
Jiatao Gu, Qi~Liu, and Kyunghyun Cho. 2019.
\newblock \href {https://doi.org/10.1162/tacl_a_00292} {Insertion-based
  decoding with automatically inferred generation order}.
\newblock \emph{Transactions of the Association for Computational Linguistics},
  7:661--676.

\bibitem[{Kingma and Ba(2015)}]{kingma15adam}
Diederik~P. Kingma and Jimmy Ba. 2015.
\newblock \href {http://arxiv.org/abs/1412.6980} {Adam: A method for stochastic
  optimization}.
\newblock In \emph{ICLR (Poster)}.

\bibitem[{Koehn et~al.(2007)Koehn, Hoang, Birch, Callison-Burch, Federico,
  Bertoldi, Cowan, Shen, Moran, Zens, Dyer, Bojar, Constantin, and
  Herbst}]{koehn-etal-2007-moses}
Philipp Koehn, Hieu Hoang, Alexandra Birch, Chris Callison-Burch, Marcello
  Federico, Nicola Bertoldi, Brooke Cowan, Wade Shen, Christine Moran, Richard
  Zens, Chris Dyer, Ond{\v{r}}ej Bojar, Alexandra Constantin, and Evan Herbst.
  2007.
\newblock \href {https://www.aclweb.org/anthology/P07-2045} {{M}oses: Open
  source toolkit for statistical machine translation}.
\newblock In \emph{Proceedings of the 45th Annual Meeting of the Association
  for Computational Linguistics Companion Volume Proceedings of the Demo and
  Poster Sessions}, pages 177--180, Prague, Czech Republic. Association for
  Computational Linguistics.

\bibitem[{Kreutzer et~al.(2020)Kreutzer, Foster, and
  Cherry}]{kreutzer-etal-2020-inference}
Julia Kreutzer, George Foster, and Colin Cherry. 2020.
\newblock \href {https://doi.org/10.18653/v1/2020.emnlp-main.465} {Inference
  strategies for machine translation with conditional masking}.
\newblock In \emph{Proceedings of the 2020 Conference on Empirical Methods in
  Natural Language Processing (EMNLP)}, pages 5774--5782, Online. Association
  for Computational Linguistics.

\bibitem[{Lample and Conneau(2019)}]{lample2019XLM}
Guillaume Lample and Alexis Conneau. 2019.
\newblock Cross-lingual language model pretraining.
\newblock \emph{Advances in Neural Information Processing Systems (NeurIPS)}.

\bibitem[{Lee et~al.(2018)Lee, Mansimov, and Cho}]{lee-etal-2018-deterministic}
Jason Lee, Elman Mansimov, and Kyunghyun Cho. 2018.
\newblock \href {https://doi.org/10.18653/v1/D18-1149} {Deterministic
  non-autoregressive neural sequence modeling by iterative refinement}.
\newblock In \emph{Proceedings of the 2018 Conference on Empirical Methods in
  Natural Language Processing}, pages 1173--1182, Brussels, Belgium.
  Association for Computational Linguistics.

\bibitem[{Lee et~al.(2020)Lee, Shu, and Cho}]{lee-etal-2020-iterative}
Jason Lee, Raphael Shu, and Kyunghyun Cho. 2020.
\newblock \href {https://doi.org/10.18653/v1/2020.emnlp-main.73} {Iterative
  refinement in the continuous space for non-autoregressive neural machine
  translation}.
\newblock In \emph{Proceedings of the 2020 Conference on Empirical Methods in
  Natural Language Processing (EMNLP)}, pages 1006--1015, Online. Association
  for Computational Linguistics.

\bibitem[{Liao et~al.(2020)Liao, Jiang, and
  Liu}]{liao-etal-2020-probabilistically}
Yi~Liao, Xin Jiang, and Qun Liu. 2020.
\newblock \href {https://doi.org/10.18653/v1/2020.acl-main.24}
  {Probabilistically masked language model capable of autoregressive generation
  in arbitrary word order}.
\newblock In \emph{Proceedings of the 58th Annual Meeting of the Association
  for Computational Linguistics}, pages 263--274, Online. Association for
  Computational Linguistics.

\bibitem[{Ma et~al.(2019)Ma, Zhou, Li, Neubig, and Hovy}]{ma-etal-2019-flowseq}
Xuezhe Ma, Chunting Zhou, Xian Li, Graham Neubig, and Eduard Hovy. 2019.
\newblock \href {https://doi.org/10.18653/v1/D19-1437} {{F}low{S}eq:
  Non-autoregressive conditional sequence generation with generative flow}.
\newblock In \emph{Proceedings of the 2019 Conference on Empirical Methods in
  Natural Language Processing and the 9th International Joint Conference on
  Natural Language Processing (EMNLP-IJCNLP)}, pages 4282--4292, Hong Kong,
  China. Association for Computational Linguistics.

\bibitem[{Mansimov et~al.(2019)Mansimov, Wang, Welleck, and
  Cho}]{mansimov2019generalized}
Elman Mansimov, Alex Wang, Sean Welleck, and Kyunghyun Cho. 2019.
\newblock A generalized framework of sequence generation with application to
  undirected sequence models.
\newblock \emph{arXiv preprint arXiv:1905.12790}.

\bibitem[{Papineni et~al.(2002)Papineni, Roukos, Ward, and
  Zhu}]{papineni-etal-2002-bleu}
Kishore Papineni, Salim Roukos, Todd Ward, and Wei-Jing Zhu. 2002.
\newblock \href {https://doi.org/10.3115/1073083.1073135} {{B}leu: a method for
  automatic evaluation of machine translation}.
\newblock In \emph{Proceedings of the 40th Annual Meeting of the Association
  for Computational Linguistics}, pages 311--318, Philadelphia, Pennsylvania,
  USA. Association for Computational Linguistics.

\bibitem[{Schulman et~al.(2017)Schulman, Wolski, Dhariwal, Radford, and
  Klimov}]{schulman2017proximal}
John Schulman, Filip Wolski, Prafulla Dhariwal, Alec Radford, and Oleg Klimov.
  2017.
\newblock Proximal policy optimization algorithms.
\newblock \emph{arXiv preprint arXiv:1707.06347}.

\bibitem[{Sennrich et~al.(2016)Sennrich, Haddow, and
  Birch}]{sennrich-etal-2016-neural}
Rico Sennrich, Barry Haddow, and Alexandra Birch. 2016.
\newblock \href {https://doi.org/10.18653/v1/P16-1162} {Neural machine
  translation of rare words with subword units}.
\newblock In \emph{Proceedings of the 54th Annual Meeting of the Association
  for Computational Linguistics (Volume 1: Long Papers)}, pages 1715--1725,
  Berlin, Germany. Association for Computational Linguistics.

\bibitem[{Shu et~al.(2020)Shu, Lee, Nakayama, and Cho}]{shu2020latent}
Raphael Shu, Jason Lee, Hideki Nakayama, and Kyunghyun Cho. 2020.
\newblock Latent-variable non-autoregressive neural machine translation with
  deterministic inference using a delta posterior.
\newblock In \emph{Proceedings of the AAAI Conference on Artificial
  Intelligence}, volume~34, pages 8846--8853.

\bibitem[{Stern et~al.(2019)Stern, Chan, Kiros, and
  Uszkoreit}]{pmlr-v97-stern19a}
Mitchell Stern, William Chan, Jamie Kiros, and Jakob Uszkoreit. 2019.
\newblock \href {http://proceedings.mlr.press/v97/stern19a.html} {Insertion
  transformer: Flexible sequence generation via insertion operations}.
\newblock In \emph{Proceedings of the 36th International Conference on Machine
  Learning}, volume~97 of \emph{Proceedings of Machine Learning Research},
  pages 5976--5985. PMLR.

\bibitem[{Vaswani et~al.(2017)Vaswani, Shazeer, Parmar, Uszkoreit, Jones,
  Gomez, Kaiser, and Polosukhin}]{vaswani2017attention}
Ashish Vaswani, Noam Shazeer, Niki Parmar, Jakob Uszkoreit, Llion Jones,
  Aidan~N Gomez, {\L}ukasz Kaiser, and Illia Polosukhin. 2017.
\newblock Attention is all you need.
\newblock In \emph{Advances in neural information processing systems}, pages
  5998--6008.

\bibitem[{Wang and Cho(2019)}]{wang-cho-2019-bert}
Alex Wang and Kyunghyun Cho. 2019.
\newblock \href {https://doi.org/10.18653/v1/W19-2304} {{BERT} has a mouth, and
  it must speak: {BERT} as a {M}arkov random field language model}.
\newblock In \emph{Proceedings of the Workshop on Methods for Optimizing and
  Evaluating Neural Language Generation}, pages 30--36, Minneapolis, Minnesota.
  Association for Computational Linguistics.

\bibitem[{Wang et~al.(2018)Wang, Zhang, and
  Chen}]{wang-etal-2018-semi-autoregressive}
Chunqi Wang, Ji~Zhang, and Haiqing Chen. 2018.
\newblock \href {https://doi.org/10.18653/v1/D18-1044} {Semi-autoregressive
  neural machine translation}.
\newblock In \emph{Proceedings of the 2018 Conference on Empirical Methods in
  Natural Language Processing}, pages 479--488, Brussels, Belgium. Association
  for Computational Linguistics.

\bibitem[{Zhang et~al.(2020)Zhang, Wang, Li, Gan, Brockett, and
  Dolan}]{zhang-etal-2020-pointer}
Yizhe Zhang, Guoyin Wang, Chunyuan Li, Zhe Gan, Chris Brockett, and Bill Dolan.
  2020.
\newblock \href {https://doi.org/10.18653/v1/2020.emnlp-main.698} {{POINTER}:
  Constrained progressive text generation via insertion-based generative
  pre-training}.
\newblock In \emph{Proceedings of the 2020 Conference on Empirical Methods in
  Natural Language Processing (EMNLP)}, pages 8649--8670, Online. Association
  for Computational Linguistics.

\end{thebibliography}
\bibliographystyle{acl_natbib}

\appendix
\section*{Appendix}

\section{Experimental Setup}

\paragraph{Datasets.} We train and evaluate our order policy on WMT'14 English-German and WMT'16 English-Romanian translation datasets.
The former contains 4.5M sentence pairs and the latter contains 2.8M sentence pairs.
We follow the preprocessing steps in previous works~\cite{lample2019XLM,mansimov2019generalized} to first tokenize each sentence using the Moses~\cite{koehn-etal-2007-moses} tokenizer and then segment each word into BPE~\cite{sennrich-etal-2016-neural} subword tokens.
For the English-German task, we use the newstest-2013 and newstest-2014 as the dev and test sets.
For the English-Romanian task, we use the newsdev-2016 and newstest-2016 as the dev and test sets.

\paragraph{De-En and En-Ro Translation Models.}
Our core translation model has the same setup as those of previous works~\cite{lample2019XLM,mansimov2019generalized}.
Specifically, we use a single stack of Transformer~\cite{vaswani2017attention} layers with 1024 hidden units, 6 layers, and 8 heads per layer.
The model is first pretrained~\cite{lample2019XLM} using a masked language modeling objective on 5M monolingual sentences from WMT NewsCrawl 2007-2008.
The model is further finetuned by~\citet{mansimov2019generalized} with a masked translation objective, where a pair of parallel English and German sentences are concatenated.
Then, tokens from a random language are masked uniformly with a ratio varying from 0\%-100\% and the model is supervised to predict the masked tokens.
We evaluate the quality of the translations using the BLEU-4~\cite{papineni-etal-2002-bleu} metric, following the setup of~\citet{ghazvininejad-etal-2019-mask} and~\citet{mansimov2019generalized}.

\paragraph{En-Zh Translation Model.}
Because the XLM repo\footnote{\url{https://github.com/facebookresearch/XLM}} only provided bilingual masked language model on De$\leftrightarrow$En, En$\leftrightarrow$Fr, and En$\leftrightarrow$Ro translation tasks, we use their multi-lingual masked language model pretrained on monolingual data in 17 languages (en-fr-es-de-it-pt-nl-sv-pl-ru-ar-tr-zh-ja-ko-hi-vi).
This transformer model has 16 layers, 8 heads per layer, and 1280 hidden units.
We then follow the same procedure as training our De$\leftrightarrow$En and En$\leftrightarrow$Ro models, first finetuning it with a masked translation objective using parallel English and Chinese sentences and then training the order policy with reinforcement learning.

\paragraph{Training Details.}
We train the order policy using Adam optimizer~\cite{kingma15adam} with a constant learning rate of $10^{-4}$, $\beta_1=0.9$, $\beta_2=0.98$.
It is trained on 4 Nvidia V100 GPUs for $\sim$72 hours with the batch size of 32 per GPU and fp16. 
Other training hyperparameters were selected based on BLEU score on the dev set with grid search: discount factor $\gamma \in \{0.9, 0.99, 0.999\}$ and the negative entropy coefficient $\lambda \in \{0.01, 0.001, 0.0005, 0.0001\}$.
We end up using $\gamma=0.999$ and $\lambda=0.001$ for all models.

\paragraph{The Maximum Decoding Length}
When we started this project, we found that the cost for running our model on the full training set is too high: using the full training set (as in~\tabref{table:full_test}) requires 4 V100 GPUs running for > 6 days. 
To select the length threshold N, we first calculate the average target sequence length for De$\xrightarrow{}$En (24.21) and Ro$\xrightarrow{}$En (27.62). 
Therefore we select an above-average threshold value of 30 that would cover more than 2/3 of the De$\xrightarrow{}$En task and only takes 3 days to train. 
All our model development/tuning are performed on the shorter subsets of the dataset.

\end{document}